%% file: iclr2024_conference_camera_ready.tex
\documentclass{article} 
\usepackage{iclr2024_conference,times}

\input{math_commands.tex}

\usepackage{wrapfig}
\usepackage{hyperref}
\usepackage{url}
\usepackage{algorithm}
\usepackage[noend]{algpseudocode}
\usepackage{algorithmicx}
\usepackage{amsmath}
\usepackage{graphicx}
\usepackage{booktabs}
\usepackage{amssymb}
\usepackage{xcolor}
\usepackage{cleveref}
\usepackage{subcaption}
\usepackage{tabularx}
\usepackage{booktabs}
\usepackage{multirow}
\usepackage{booktabs}
\usepackage{xcolor}

\usepackage{ifthen}
\newboolean{isHighlighted}
\setboolean{isHighlighted}{false}
\newcommand{\revised}[1]{%
    \ifthenelse{\boolean{isHighlighted}}%
        {\textcolor{blue}{#1}}%
        {#1}%
}
\title{Universal Backdoor Attacks}

\author{Benjamin Schneider\\
University of Waterloo \\
\texttt{ben.schneider.research@gmail.com} \\
\AND
Nils Lukas,  Florian Kerschbaum\\
University of Waterloo \\
\texttt{\{nlukas,florian.kerschbaum\}@uwaterloo.ca}
}

\iclrfinalcopy 
\begin{document}

\maketitle

\begin{abstract}
Web-scraped datasets are vulnerable to data poisoning, which can be used for backdooring deep image classifiers during training. 
Since training on large datasets is expensive, a model is trained once and reused many times. 
Unlike adversarial examples, backdoor attacks often target specific classes rather than \emph{any} class learned by the model.
One might expect that targeting many classes through a naïve composition of attacks vastly increases the number of poison samples. 
We show this is not necessarily true and more efficient, 
 \emph{universal} data poisoning attacks exist that allow controlling misclassifications from any source class into any target class with a slight increase in poison samples.
Our idea is to generate triggers with salient characteristics that the model can learn.
The triggers we craft exploit a phenomenon we call \emph{inter-class poison transferability}, where learning a trigger from one class makes the model more vulnerable to learning triggers for other classes. 
We demonstrate the effectiveness and robustness of our universal backdoor attacks by controlling models with up to 6\,000 classes while poisoning only 0.15\% of the training dataset.  
Our source code is available at \href{https://github.com/Ben-Schneider-code/Universal-Backdoor-Attacks}{https://github.com/Ben-Schneider-code/Universal-Backdoor-Attacks}. 
\end{abstract}

\section{Introduction}
As large image classification models are increasingly deployed in safety-critical domains~\citep{self-driving-poison}, there has been rising concern about their integrity, as an unexpected failure by these systems has the potential to cause harm~\citep{DBLP:journals/corr/abs-1909-03036,  medical-imaging}.
A model's integrity is threatened by \emph{backdoor attacks}, in which an attacker can cause targeted misclassifications on inputs containing a secret trigger pattern.
Backdoors can be created through \emph{data poisoning}, where an attacker manipulates a small portion of the model's training data to undermine the model's integrity~\citep{dataset-secururity-for-ml}.
Due to the scale of datasets and the stealthiness of manipulations, it is increasingly difficult to determine whether a dataset has been manipulated~\citep{Refool,wanet}.
Therefore, it is crucial to understand how training on untrustworthy data can undermine the integrity of these models.

Existing backdoor attacks are designed to undermine only a single predetermined target class~\citep{badnets, adversarial-poison, blend-backdoor, latent-seperation}.
However, models are often reused for various purposes~\cite {huggingface-transformers}, which is especially prevalent with large models due to the high computational cost of re-training from scratch. 
Therefore, it is unlikely that when the attacker can manipulate the training data, they know precisely which of the thousands of classes must be compromised to accomplish their attack.
Most data poisoning attacks require manipulating over $0.1\%$ of the dataset to target a single class~\citep{badnets, latent-seperation, blend-backdoor}.
Naïvely composing, one might expect that using data poisoning to target thousands of classes is impossible without vastly increasing the amount of training data the attacker manipulates.
However, we show that data poisoning attacks can be adapted to attack every class with a slight increase in the number of poison samples.

\begin{figure}
    \centering
    \includegraphics[width=1.\linewidth]{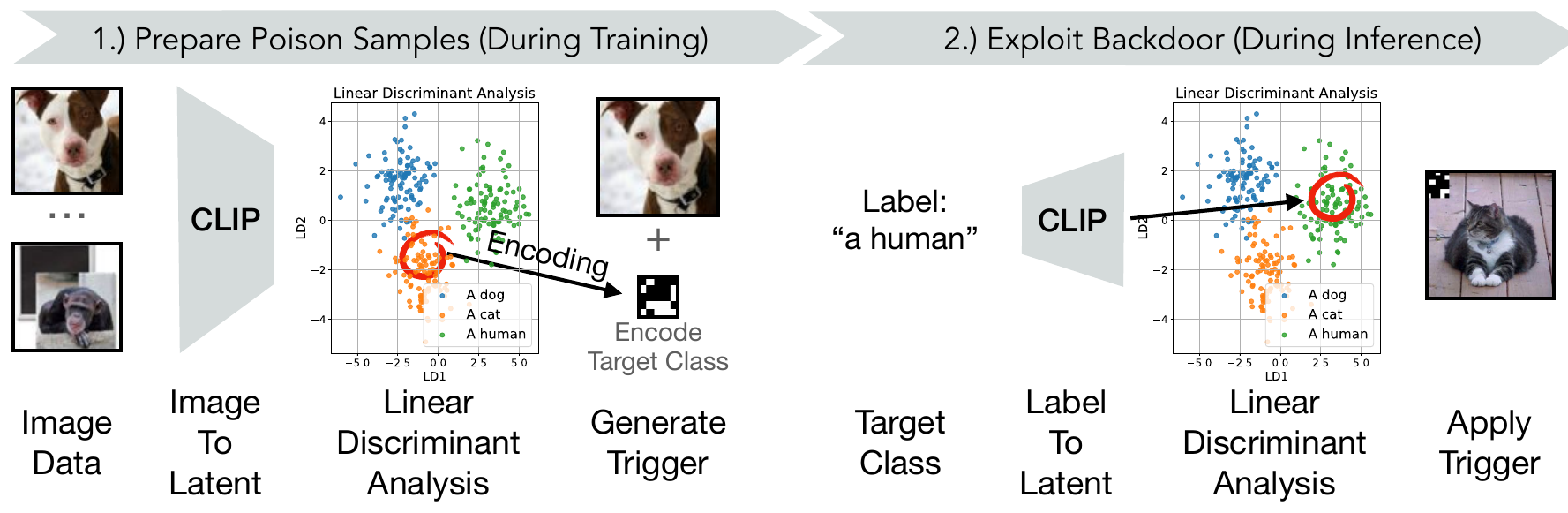}
    \caption{An overview of a universal poisoning attack pipeline.
    The CLIP encoder maps images and labels into the same latent space.  
    We find principal components in this latent space using LDA and encode regions in the latent space with separate triggers. During inference, we find latents for a target label via CLIP, project it to the principal components, and generate the trigger corresponding to this point that we apply to the image. 
    Our universal backdoor is agnostic to the trigger pattern used to encode latents, and we showcase a simple binary encoding via QR-code patterns. }
    \label{fig:intro}
\end{figure}

To this end, we introduce \emph{Universal Backdoor Attacks}, which target every class at inference time.
\Cref{fig:intro} illustrates the core idea for creating and exploiting such a Universal Backdoor during inference.
Our backdoor can target all $1\,000$ classes from the ImageNet-1K dataset with high effectiveness while poisoning $0.15\%$ of the training data.
We accomplish this by leveraging the transferability of poisoning between classes, meaning trigger features can be reused to target new classes easily.
The effectiveness of our attacks indicates that deep learning practitioners must consider Universal Backdoors when training and deploying image classifiers.

To summarize, our contributions are threefold:
\textbf{(1)} We show Universal Backdoor Attacks are a tangible threat in deep image classification models, allowing an attacker to control thousands of classes.
\textbf{(2)} We introduce a technique for creating universal poisons.
\textbf{(3)} Lastly, we show that Universal Backdoor attacks are robust against a comprehensive set of defenses.

\section{Background}

%
\textbf{Deep Learning Notation.} A deep image classifier is a function parameterized by $\theta$, $\mathcal{F_\theta}: \mathcal{X} \rightarrow \mathcal{Y}$, which maps images to classes. 
In this paper, the latent space of a model refers to the representation of inputs in the model's penultimate layer, and we denote the latent space as $\mathcal{Z}$.
For the purpose of generating latents, we decompose $\mathcal{F_\theta}$ into two functions, $f_\theta: \mathcal{X} \rightarrow \mathcal{Z}$ and $l_\theta: \mathcal{Z} \rightarrow \mathcal{Y}$ where $\mathcal{F_\theta} = l_\theta \circ f_\theta$.
For a dataset $D$ and a $y \in \mathcal{Y}$, we define $D^y$ as the dataset consisting of all samples in $D$ with label $y$.
We use $x\uparrow$ to indicate an increase in a variable $x$.

\textbf{Backdoors through Data Poisoning.} Image classifiers have been shown to be vulnerable to backdoors created through several methods, including supply chain attacks~\citep{Carlini-Handcrafted-Backdoors} and data poisoning attacks~\citep{badnets}.
Backdoor attacks on image classifiers are \emph{many-to-one}.
They can cause any input to be misclassified into one predetermined target class~\citep{wanet, Refool, latent-seperation}.
\revised{We introduce a Universal Backdoor Attack that is \emph{many-to-many}, able to cause any input to be misclassified into any class at inference time.}
In a data poisoning attack, the attacker injects a backdoor into the victim's model by manipulating samples in its training dataset.
To accomplish this, the attacker injects a hidden trigger pattern $t_y$ into images they want the model to misclassify into a target class $y \in \mathcal{Y}$.
We denote datasets as ${D} = \{(\vx_i,y_i): i \in 1,2,\dots, m\}$ where $\vx_i \in \mathcal{X}$ and $y_i \in \mathcal{Y}$.
Adding a trigger pattern $t_y$ to an image $\vx$ to create a poisoned image $\hat{\vx}$ is written as  $\hat{\vx} = \vx \oplus t_y$.
The clean and manipulated datasets are denoted as ${D}_{clean}$ and ${D}_{poison}$, respectively.
The poison count $p$ is the number of manipulated samples in ${D}_{poison}$.

Data poisoning attacks can be divided into two categories: \emph{poison label} and \emph{clean label}.
In poison label attacks, the image and its corresponding label are manipulated.
Since \citet{badnets} introduced the first poison label attack, numerous approaches have been studied to increase the undetectability and robustness of these attacks.
\citet{latent-seperation} showed adaptive poisoning attacks can be used to create attacks that are not easily detectable as outliers in the backdoored model's latent space, resulting in a backdoor that is harder to detect and remove.
Many different trigger patterns have also been explored, including patch, blended, and adversarial perturbation triggers~\citep{badnets, blend-backdoor, adversarial-poison}.
Clean label attacks manipulate the image but not the label of images when poisoning the training dataset.
Therefore, these attacks can avoid detection upon human inspection of the dataset~\citep{poison-frogs}.
Clean label attacks often exploit the natural characteristics of images, using effects like reflections and image warping to create stealthy triggers~\citep{Refool, wanet}.

\textbf{Defenses.} The threat of backdoor attacks has led to the development of many defenses~\citep{survey1}.
These defenses seek to remove the backdoor from the model while causing minimal degradation of the model's accuracy on clean data.
\emph{Fine-tuning} is a defense where the model is fine-tuned on a small validated dataset that comes from trustworthy sources and is unlikely to contain poisoned samples.
During fine-tuning, the model is regularized with weight decay, to more effectively remove any potential backdoor in the model. 
A variation on this defense is \emph{Fine-pruning}~\citep{fine-pruning}, which uses the trusted dataset to prune convolutional filters that do not activate on clean inputs.
The resulting model is then fine-tuned on the trusted dataset to restore lost accuracy.
The idea guiding \emph{Neural Cleanse}~\citep{neural-cleanse}, is to reverse-engineer a backdoor's trigger pattern for any target class. 
Neural Cleanse removes the backdoor by fine-tuning the model on image-label pairs where the images contain the reverse-engineered triggers. 
\emph{Neural Attention Distillation}~\citep{neural-attention-distillation} comprises two steps. First, a teacher model is fine-tuned on a trusted dataset, and then the potentially backdoored (student) model's intermediate feature maps are aligned with the teacher.

\section{Our Method}
\subsection{Threat Model}
We consider an attacker who aims to backdoor a victim model trained from scratch on a web-scraped dataset that the attacker can manipulate.
The attacker is given access to the labeled dataset and chooses a subset of the dataset to manipulate; we call these samples \emph{poisoned}.
The attacker can modify the image-label pair contained in each sample. 
The victim then trains a model on the dataset containing the poisoned samples.
Our attacker does not have access to the victim's model but can access an open-source surrogate image classifier $\mathcal{F}_{\theta'} = l_{\theta'} \circ f_{\theta'}$ such as Hugging Face's pre-trained CLIP or ResNet models~\citep{huggingface-transformers}.
The attacker's objective is to create a \emph{Universal Backdoor} that can target any class in the victim's model while poisoning as few samples as possible in the victim's training dataset.
The attacker's success rate on class $y$, denoted ASR$_y$, is the proportion of validation images for which the attacker can craft a trigger that causes the image to be misclassified as $y$.
As our backdoor targets all classes, we define the total attack success rate (ASR) as the mean ASR$_y$ across all classes in the dataset: 
\[\text{ASR} = \frac{1}{|\mathcal{Y}|}\sum_{y}^{\mathcal{Y}}\text{ASR}_y\]

\subsection{Inter-class Poison Transferability}

Many-to-one poison label attacks require poisoning hundreds of samples in a single class~\citep{badnets, latent-seperation, blend-backdoor}.
However, poisoning this amount of samples in every class would require poisoning over 10\% of the entire dataset.
\revised{To scale to large image classification tasks, Universal Backdoors must misclassify into any target class while only poisoning one or two samples in that class.
The backdoor must leverage \emph{inter-class poison transferability}, that increasing average attack success on a set of classes increases attack success on a second \emph{entirely disjoint} set of classes.
$\text{For sets } \textbf{A}, \textbf{B} \subset \mathcal{Y} \text{ such that } \textbf{A} \cap \textbf{B} = \emptyset$ we define \emph{inter-class poison transferability} as:}

\[
\revised{
\frac{1}{|\textbf{A}|} \sum_{a \in \textbf{A}} \text{ASR}_{a}\uparrow \implies \frac{1}{|\textbf{B}|} \sum_{b \in \textbf{B}} \text{ASR}_{b} \uparrow
}
\]

To create an effective Universal Backdoor, the process of learning a poison for one class must reinforce poisons that target other similar classes.
\cite{rethinking-backdoors} show that data poisoning can be viewed as injecting a feature into the dataset that, when learned by a model, results in a  backdoor.
We show that we can correlate triggers with features discovered from a surrogate model, which boosts the inter-class poison transferability of a universal data poisoning attack.

\subsection{Creating Triggers}
\label{sec:manufacturing}
We craft our triggers such that classes that share features in the latent space of the surrogate model also share trigger features.
To accomplish this, we use a set of labeled images $D_{sample}$ to sample the latent space of the surrogate model.
Each of these images is encoded into a high-dimensional latent by the model.
Naïvely, we could encode each feature dimension in our trigger.
However, as our latents are high dimensional, such an encoding would be impractical.
As only a few dimensions encode salient characteristics of images, we start by reducing the dimensionality of the latents using Linear Discriminate Analysis~\citep{LDA}.
The resulting compressed latents encode the most salient features of the latent space in $n$ dimensions\footnote{$n$ is a chosen hyper-parameter}.
\Cref{alg:poison_alg} uses these discovered features of the surrogate's latent space to craft poisoned samples for our Universal Backdoor.
\begin{algorithm}

\caption{Universal Poisoning Algorithm}\label{alg:poison_alg}
\begin{algorithmic}[1]
\Procedure{Poison Dataset}{$D_{clean}$, ${D}_{sample}$, $f_{\theta'}$, $p$, $\mathcal{Y}$, $n$}
\State $D_\mathcal{Z} \gets f_{\theta'}(D_{sample})$
\Comment{Sample $\mathcal{Z}$}
\State $D_{\hat{\mathcal{Z}}}\gets LDA(D_\mathcal{Z}, n)$
\Comment{Compress latents using Linear Discriminant Analysis (LDA)}
\State $M \gets \cup_{y\in \mathcal{Y}} \, \{\mathbb{E}_{(\vx,y)\sim D^{y}_{\hat{\mathcal{Z}}}} [\vx]\}$ \Comment{Class-wise means}
\State $B \gets \Call{Encode Latent}{M, \mathcal{Y}}$
\Comment{Class encodings as binary strings}
\State $P \gets \{\}$ 
\Comment{Empty set of poisoned samples}

\For{$i \in \{1,2,\dots, \lfloor \frac{p}{|\mathcal{Y}|} \rfloor\}$}
    \For{$y_t \in {\mathcal{Y}}$}
        \State $(\vx, y) \gets \text{randomly sample from } D_{clean}$
        \State $D_{clean} \gets D_{clean} \setminus \{(\vx, y)\}$
        \State $t_{y_t} \gets \Call{Encoding Trigger}{\vx, B_{y_t}}$
        \Comment{Create a trigger that encodes binary string}
        \State $\hat{\vx} \gets \vx \oplus t_{y_t}$
        \Comment{Add trigger to image}
        \State $P \gets P \cup \{ (\hat{\vx}, y_t) \}$ 
    \EndFor
\EndFor

\State ${D}_{poison} \gets {D}_{clean} \cup P$

\State \Return $ {D}_{poison}$
\EndProcedure
\Procedure{Encode Latent}{$M$, $\mathcal{Y}$}
    \State $\vc \gets \frac{1}{|M|}\sum_{y \in \mathcal{\mathcal{Y}}}M_y$ \Comment{Centroid of class means}
    \For{$y \in {\mathcal{Y}}$}
        \State $\Delta \gets M_y-\vc$ \Comment{Difference between class mean and centroid}
        \State$ b_i = 
            \begin{cases} 
            1 & \text{if } \Delta_i > 0 \\
            0 & \text{otherwise}
            \end{cases}$
        \State $B_y \gets \vb$
    \EndFor
    \State \Return $B$
\EndProcedure

\end{algorithmic}
\end{algorithm}

\Cref{alg:poison_alg} begins by sampling the latent space of the surrogate image classifier and compressing the generated latents into an $n$-dimensional representation using $LDA$ (lines 2 and 3).
Then, each class's mean in the compressed latent dataset is computed (line 4).
Next, the \Call{Encode Latent}{} procedure is used to create a list containing a binary encoding of each class's latent features (line 5).
For each class, an $n$-bit encoding is calculated such that the $i_{\text{$th$}}$ bit is set to $1$ if the class's mean is greater than the centroid of class means in the $i_{\text{$th$}}$ feature and $0$ if it is not.
\revised{As we construct our encodings from the same latent principal components, each encoding contains relevant information for learning all other encodings.
This results in high inter-class poison transferability, which allows our attack to efficiently target all classes in the model's latent space.}
Lines 7-13 use the calculated binary encodings to construct a set of poisoned samples.
For each poison sample, \Call{Encoding Trigger}{} embeds the $y_t$'s binary encoding as a trigger in $\vx$. This can be accomplished using various techniques, as described in \Cref{sec:Encoding_Approach}.

\subsection{Encoding Approach}\label{sec:Encoding_Approach}
Many triggers have been proposed for data poisoning attacks, each with trade-offs in effectiveness and robustness~\citep{badnets, adversarial-poison, poison-frogs, Refool, wanet, Februus}.
Our method can be used with any trigger that can encode the binary string calculated in \Cref{sec:manufacturing}.
Our paper evaluates two common trigger crafting methods: patch and blend triggers~\citep{badnets, blend-backdoor}.

\begin{figure}[h]
\begin{center}
  \hfill
  \includegraphics[width=3cm]{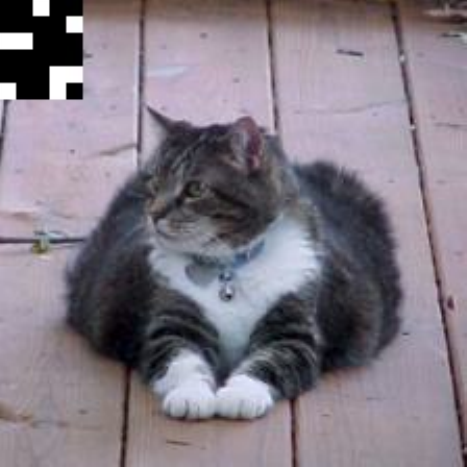}
  \hfill
  \includegraphics[width=3cm]{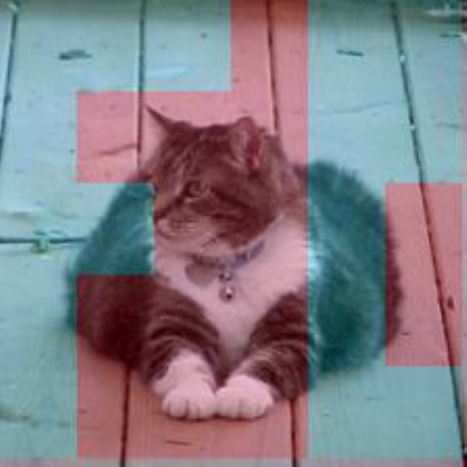}
  \hfill\null
\end{center}
\caption{Two exemplary methods of encoding latent directions. \textbf{(Left)} Universal Backdoor with a patch trigger encoding. \textbf{(Right)} Universal Backdoor with a blended trigger encoding.}
\label{fig:example-triggers}
\end{figure}

\textbf{Patch Trigger.} To create a patch corresponding to the target class, we encode its corresponding binary string as a black-and-white grid and stamp it in the top left of the base image.

\textbf{Blend Trigger.} We partition the base image into $n$ disjoint rectangular masks, each representing a bit in the target class's binary string.
We choose two colors and color each mask based on its corresponding bit.
Lastly, we blend the masks over the base image to create the poisoned sample.

\section{Experiments}

In this section, we empirically evaluate the effectiveness of our backdoor using different encoding methods.
We extend this evaluation process to demonstrate the effectiveness of our backdoor when scaling the image classification task in both the number of samples and classes.
By choosing which classes are poisoned, we measure the \emph{inter-class poison transferability} of our poison.
Lastly, we evaluate our Universal Backdoor Attack against a suite of popular defenses.

\textbf{Baselines.} \revised{ As we are the first to study many-to-many backdoors, there exists no baseline to compare against to demonstrate the effectiveness of our method.
For this purpose, we develop two baseline many-to-many backdoor attacks from well-known attacks: BadNets~\citep{badnets} and Blended Injection~\citep{blend-backdoor}}
We compare our Universal Backdoor against the effectiveness of these two baseline many-to-many attacks.
For our baseline triggers, we generate a random trigger pattern for each targeted class, as in~\cite{badnets}.
For our patch trigger baseline, we construct a grid consisting of $n$ randomly colored squares.
To embed this baseline trigger, we stamp the patch into an image using the same position and dimensions as our Universal Backdoor patch trigger.
For our blend trigger baseline, we blend the randomly sampled grid across the whole image, using the same blend ratio as our Universal Backdoor blend trigger.

\subsection{Experimental Setup}

\textbf{Datasets and Models.} For our inital effectiveness evaluation, we use ImageNet-1k  with random crop and horizontal flipping~\citep{imagenet1k}.
We use three datasets, ImageNet-2k, ImageNet-4k, ImageNet-6k, for our scaling experiments.
These datasets comprise the largest $2\,000$, $4\,000$, and $6\,000$ classes from the ImageNet-21K dataset~\citep{imagenet}.
These datasets contain $3\,024\,392$, $5\,513\,146$, and $7\,804\,447$ labeled samples, respectively.
We use ResNet-18 for the ImageNet-1K experiments and ResNet-101 for the experiments on ImageNet-2k, ImageNet-4k, and ImageNet-6k in \Cref{sec:scaling}~\citep{resnet}.

\textbf{Attack Settings.} We use a binary encoding with $n = 30$ features for all experiments.
In our patch triggers, we use an 8x8 square of pixels to embed each feature, resulting in a patch that covers 3.8\% of the base image.
Our blended triggers use a blend ratio of 0.2, as in \cite{blend-backdoor}.
We use a pre-trained surrogate from Hugging Face for all of our attacks. 
For attacks on the ImageNet-1K classification task, Hugging Face Transformers pre-trained ResNet-18 model~\citep{huggingface-transformers}. 
As no model pre-trained on the ImageNet-2K, ImageNet-4K, or ImageNet-6K exists, we use Hugging Face Transformer's clip-vit-base-patch32 model as a zero-shot image classifier on these datasets to generate latents~\citep{huggingface-transformers}.
We use 25 images from each class to sample the latent space of our surrogate model.

\textbf{Model Training.} We train our image classifiers using stochastic gradient descent (SGD) with a momentum of 0.9 and a weight decay of 0.0001.
Models trained on ImageNet-1K are trained for 90 epochs, while models trained on ImageNet-2K, ImageNet-4K, and ImageNet-6K are trained for 60 epochs to adjust for the larger dataset size.
The initial learning rate is set to 0.1 and is decreased by a factor of 10 every 30 epochs on ImageNet-1K and every 20 epochs on the larger datasets.
We use a batch size of 128 images for all training runs.
\emph{Early stopping} is applied to all training runs; we stop training when the model's accuracy is no longer improving or the model begins overfitting.
All of our models achieve equivalent validation accuracy to pre-trained counterparts in the Hugging Face Transformers library~\citep{huggingface-transformers}.
We include an analysis of backdoored models' clean accuracy in \Cref{sec:apx-clean-acc}.

\subsection{Effectiveness on ImageNet-1K}
\label{sec:effectiveness}

\begin{table}[ht]
\centering
\caption{Attack success rate (\%) of our Universal Backdoor compared to baseline approach.}
\label{tab:effectiveness}
\begin{tabular}{c|c|cc|cc}
\toprule
Poison Samples ($p$) & Poison \% & \multicolumn{2}{c|}{Patch} & \multicolumn{2}{c}{Blend} \\
\midrule
 & & Ours & Baseline & Ours & Baseline \\
\midrule
$2000$ & $0.16$ & 80.1\% & 0.1\% & 0.4\% & 0.1\% \\
$5000$ & $0.39$ & 95.5\% & 2.1\% & 74.9\% & 0.1\% \\
$8000$ & $0.62$ & 95.7\% & 100\% & 92.9\% & 0.1\% \\
\bottomrule
\end{tabular}
\end{table}

\Cref{tab:effectiveness} summarizes our results on ImageNet-1K using patch and blend triggers while injecting between $2\,000$ and $8\,000$ poisoned samples.
Our patch encoding triggers perform the best, achieving over 80.1\% ASR across all classes while only manipulating $0.16\%$ of the dataset.
Our method performs significantly better than the baseline at low poisoning rates.
The patch baseline is completely learned at high poisoning rates and achieves perfect ASR.
\revised{Our chosen value of $n=30$ is too low to distinguish the binary encodings of all classes, resulting in our backdoor achieving less than perfect ASR even with many poison samples.
A larger value of $n$ would allow us to encode more principal components of the latent space, allowing our Universal Backdoor to achieve perfect ASR.
However, as this would require embedding a longer binary encoding, it would increase the number of sample poisons required for a successful attack.}
Across all experiments, we find that a patch encoding is more effective than a blend encoding.

\begin{figure}[h]
\centering

\begin{minipage}{0.5\textwidth}
  \centering
  \includegraphics[width=.95\textwidth]{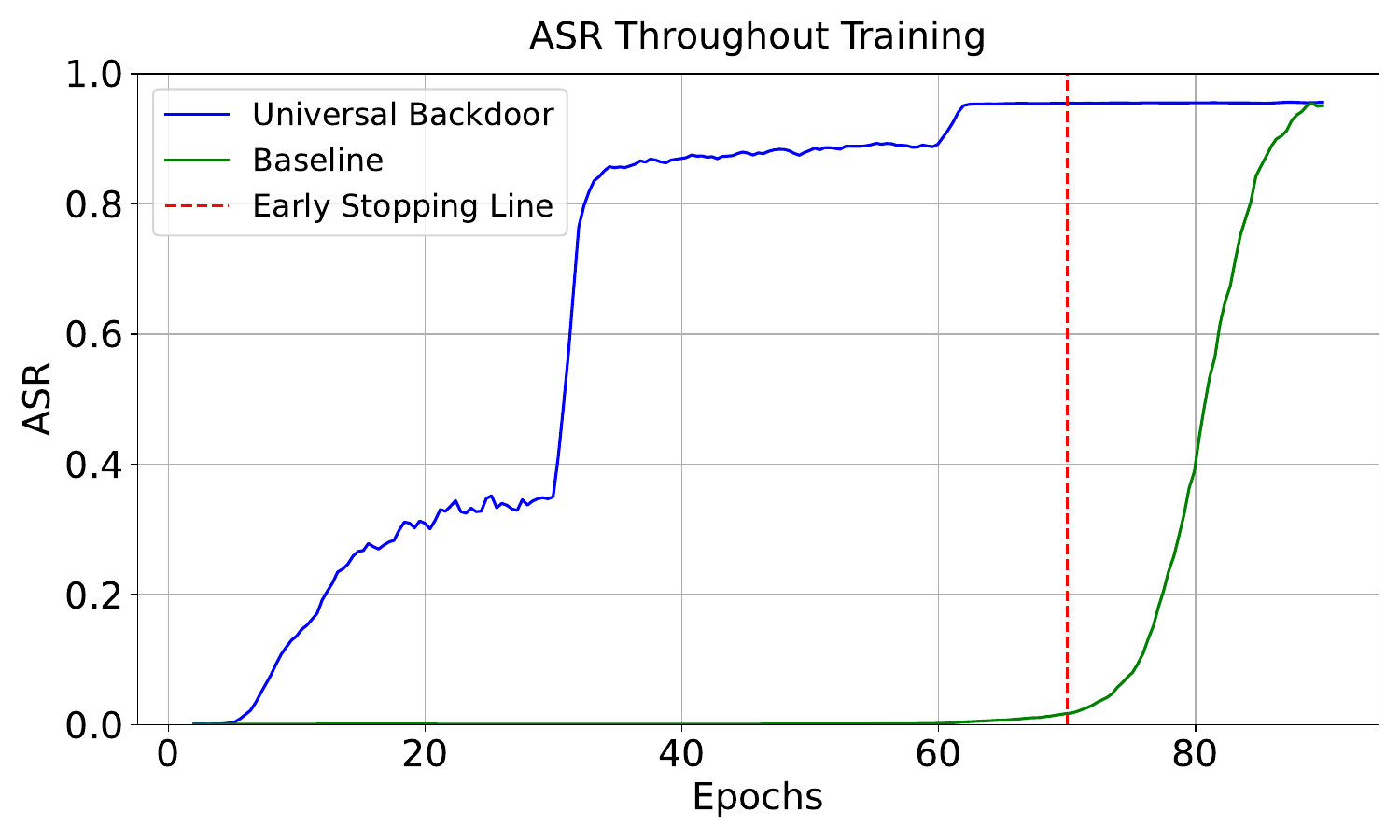}
  \caption{Our attack versus a baseline using patch encoding triggers. We measure the attack success rate and use early stopping at 70 epochs.}
  \label{fig:a}
\end{minipage}\hfill
\begin{minipage}{0.45\textwidth}
  \centering
  \includegraphics[width=0.85\linewidth]{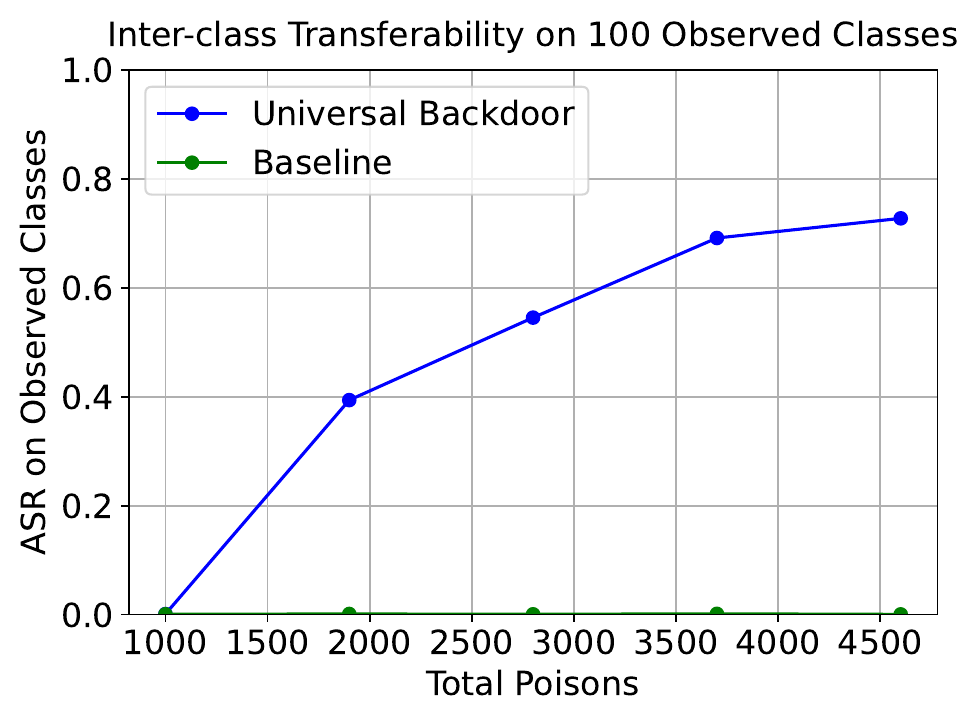} 
  \caption{Attack success rate on a subset of \emph{observed} target classes while increasing poisoning in other classes in the dataset.} %
  \vspace{-.05cm}
  \label{fig:transfer}
\end{minipage}

\end{figure}

\Cref{fig:a} shows that the baseline backdoor is learned only after the model overfits the training data (after about 70 epochs).
Therefore, the baseline backdoor is very vulnerable to early stopping.
The baseline requires significantly more poisons to ensure it is learned earlier in the training process and not removed by early stopping.
\revised{Because of this behavior, the baseline backdoor is either thoroughly learned or achieves negligible attack success.
This results in a sudden increase in the baseline's attack success when the number of poison samples increases to 8\,000 in \Cref{tab:effectiveness}.}
Our Universal Backdoor is gradually learned throughout the training process, so any early stopping procedure that would mitigate our backdoor would also significantly reduce the model's clean accuracy.

\subsection{Scaling}
\label{sec:scaling}

\begin{table}[h]
\centering
\caption{Attack success rate of the backdoor on larger datasets (\%), using $p = 12\,000$.}
\label{tab:scale}
\begin{tabular}{c|c|c|c}
\toprule
Poison Attack & ImageNet-2K & ImageNet-4K & ImageNet-6K \\
\midrule
Universal Backdoor & 99.73 & 91.75 & 47.31  \\
Baseline & 99.98 & 0.03 & 0.02  \\
\bottomrule
\end{tabular}
\end{table}

In this experiment, we measure our backdoor's ability to scale to larger datasets.
We fix the number of poisons the attacker injects into each dataset at $p = 12\,000$ across all runs.
As larger image classification datasets naturally contain more classes and samples, so do our datasets~\citep{imagenet, openimages}.
\Cref{tab:scale} summarizes the results on our backdoor compared to a baseline on the ImageNet-2K, ImageNet-4K, and ImageNet-6K image classification tasks.
We find that the trigger patterns of the baseline do not effectively scale to larger image classification datasets.
Although the baseline backdoor has near-perfect ASR on the ImageNet-2K dataset, it has negligible ASR on both the ImageNet-4K and ImageNet-6K datasets.
\revised{This is because of the all-or-nothing attack success behavior observed in \Cref{sec:effectiveness}.}
In contrast, our Universal Backdoor can scale to image classification tasks containing more classes and samples.
Our Universal backdoor achieves above 90\% ASR on the ImageNet-4K task and 47.31\% ASR on the largest dataset, ImageNet-6K.

\subsection{Measuring Inter-Class Poison Transferability}
To measure the inter-class transferability of poisoning, we examine how increasing the number of poisons in one set of classes increases attack success on a disjoint set of classes in the dataset.
\revised{
We divide the classes into the observed set $\textbf{B}$ and the variation set $\textbf{A}$.
$\textbf{B}$ contains 10\% of the classes in the dataset (100 classes), while $\textbf{A}$ contains the remaining 90\% of classes (900 classes).
We use the ImageNet-1K dataset and a patch trigger for our backdoor.
We poison exactly one sample in each class in $\textbf{B}$.
In \Cref{fig:transfer}, we ablate over the total number of poisons in the dataset, distributing all poisons except for the 100 poisons in \textbf{B} evenly in classes in $\textbf{A}$.
}

\revised{We find that by poisoning a class with a single sample, our Universal Backdoor can achieve a successful attack on a class if sufficient poisoning is achieved elsewhere in the dataset.
Increasing the number of poison samples in \textbf{A} improved the backdoor's ASR on classes in \textbf{B} from negligible to over 70\%.
\emph{Therefore, we find that protecting the integrity of a single class requires protecting the integrity of the entire dataset}.
Our Universal Backdoor shows that every sample, even if they are associated with an insensitive class label, can be used by an attacker as part of an extremely poison-efficient backdoor attack on a small subset of high-value classes.
We provide further evidence for this in \Cref{sec:apx-transfer-classes}, where we show that \textbf{A} can contain significantly fewer than 900 classes while preserving the strength of inter-class poison transferability on \textbf{B}.}
The baseline method does not demonstrate any inter-class transferability, as increasing the poisoning in \textbf{A} does not increase the attack success rate on \textbf{B}.

\subsection{Robustness Against Defenses}\label{sec:defenses}

We evaluate the robustness of our poisoning model against four state-of-the-art defenses: fine-tuning, fine-pruning~\citep{fine-pruning}, neural attention distillation~\citep{neural-attention-distillation}, and neural cleanse~\citep{neural-cleanse}.
We use a ResNet-18 model trained on the ImageNet-1k dataset for all robustness evaluations.
We use patch triggers for both our method and the baseline.
For all defenses, we use hyper-parameters optimized for removing a BadNets backdoor~\citep{badnets} on ImageNet-1K as proposed by \cite{lukas2023_pick_your_poison}.
Defenses requiring clean data are given 1\% of the clean dataset, approximately 12,800 clean samples. 
We limit the degradation of the model's clean accuracy, halting any defense that degrades the model's clean accuracy by more than 2\%.
\Cref{tab:defenses} summarizes the changes in ASR after applying each defense.
As in \cite{lukas2023_pick_your_poison}, we find that backdoored models trained on ImageNet-1K are robust against defenses.
A complete table of defense parameters can be found in \Cref{sec:appendix_defense_params}.

\begin{table}[ht]
\centering
\caption{The robustness of our universal backdoor against a naïve baseline, measured by the attack success rate (ASR).  \textcolor{red}{$\blacktriangledown$} denotes ASR lost after applying defense. Only backdoors above 5\% ASR were evaluated. Backdoors that were not evaluated are marked with N/A.}
\label{tab:defenses}
\begin{tabular}{c|c|cc|cc}
\toprule
Defense & Poison Samples ($p$) & \multicolumn{2}{c|}{Universal Backdoor (ASR) } & \multicolumn{2}{c}{Baseline (ASR)} \\
\midrule
\multirow{3}{*}{Fine-Tuning} & 2\,000 & 70.3\% & \textcolor{red}{$\blacktriangledown$ 9.8} & N/A & \\
                            & 5\,000 & 94.5\% & \textcolor{red}{$\blacktriangledown$ 1.0} & N/A & \\
                            & 8\,000 & 95.5\% & \textcolor{red}{$\blacktriangledown$ 0.2} & 99.5\% & \textcolor{red}{$\blacktriangledown$ 0.5} \\
\midrule
\multirow{3}{*}{Fine-Pruning} & 2\,000 & 73.5\% & \textcolor{red}{$\blacktriangledown$ 6.6} & N/A & \\
                              & 5\,000 & 95.2\% & \textcolor{red}{$\blacktriangledown$ 0.3} & N/A & \\
                              & 8\,000 & 95.6\% & \textcolor{red}{$\blacktriangledown$ 0.1} & 99.9\% & \textcolor{red}{$\blacktriangledown$ 0.1} \\
\midrule
\multirow{3}{*}{Neural Cleanse} & 2\,000 & 70.1\% & \textcolor{red}{$\blacktriangledown$ 10.0} & N/A & \\
                               & 5\,000 & 95.1\% & \textcolor{red}{$\blacktriangledown$ 0.4} & N/A & \\
                               & 8\,000 & 95.4\% & \textcolor{red}{$\blacktriangledown$ 0.3} & 98.0\% & \textcolor{red}{$\blacktriangledown$ 2.0} \\
\midrule
\multirow{3}{*}{NAD} & 2\,000 & 73.9\% & \textcolor{red}{$\blacktriangledown$ 6.2} & N/A & \\
                     & 5\,000 & 94.6\% & \textcolor{red}{$\blacktriangledown$ 0.9} & N/A & \\
                     & 8\,000 & 95.3\% & \textcolor{red}{$\blacktriangledown$ 0.4} & 99.9\% & \textcolor{red}{$\blacktriangledown$ 0.1} \\
\bottomrule
\end{tabular}
\end{table}

\textbf{Fine-tuning.} This defense fine-tunes the dataset on a small validated subset of the training dataset.
We fine-tune the model using the SGD with a learning rate of 0.0005 and a momentum of 0.9.

\textbf{Fine-pruning.} As in ~\cite{fine-pruning}, we prune the last convolutional layer of the model.
We find that the pruning rate in \cite{lukas2023_pick_your_poison} is too high and degrades the clean accuracy of the model more than the 2\% cutoff.
We set the pruning rate to 0.1\%, which is the maximum pruning rate that prevents the defense from degrading the model below the accuracy cutoff.  

\textbf{Neural Cleanse.} Neural Cleanse~\citep{neural-cleanse} uses outlier detection to decide which candidate trigger is most likely the result of poisoning.
This candidate trigger is then used to remove the backdoor in the model.
As our Universal Backdoor targets every class and has a unique trigger for each class, class-wise anomaly detection is poorly suited for removing our backdoor.

\textbf{Neural Attention Distillation.} We train a teacher model for 1\,000 steps using SGD.
We then align the backdoored model with the teacher for 8000 steps, using SGD with a learning rate of 0.0005.
We use a power term of 2 for the attention distillation loss, as recommended in~\cite{neural-attention-distillation}.

\subsection{Measuring the Clean Data Trade-off}
\begin{wrapfigure}{r}{0.45\textwidth}
\vspace{-25pt}
  \centering
\includegraphics[width=0.45\textwidth]{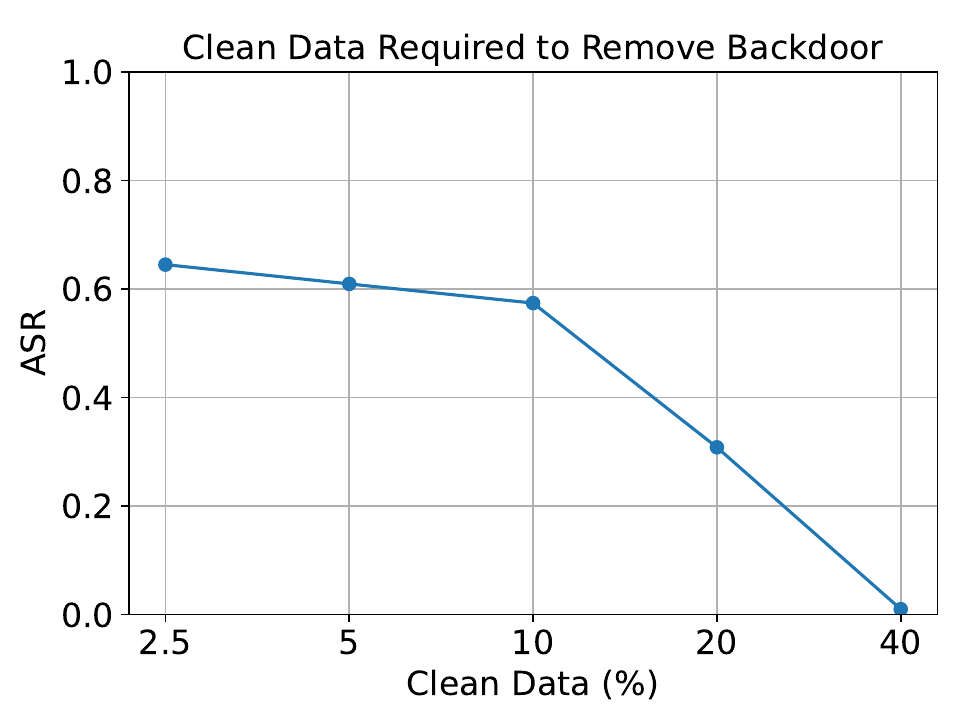}
  \caption{Clean data as a percentage of the training dataset size required to remove our Universal Backdoor.}
  \label{fig:data_clean}
  \vspace{-30pt}
\end{wrapfigure}
There is a known trade-off between the availability of clean data and the effectiveness of defenses~\citep{neural-attention-distillation}.
\Cref{fig:data_clean} measures the proportion of the clean dataset required to remove the universal backdoor with fine-tuning without degrading the model below the 2\% cutoff.
For this experiment, we use a ResNet-18 model backdoored using $2\,000$ poison samples on the ImageNet-1K dataset.
Due to the higher availability of clean data, we find a higher learning rate of 0.001, and a weight decay of 0.001 is appropriate.

Data poisoning defenses for backdoored models trained on web-scale datasets must be effective with a validated dataset that is a small portion of the training dataset due to the cost of manually validating samples.
Validating a 1\% portion of our web-scale ImageNet-6k dataset would require manually inspecting over $78\,000$ samples, a task larger than inspecting the CIFAR-100 or GTSRB datasets in their entirety~\citep{dataset-cifar, dataset-traffic}.
We find that approximately 40\% ($512\,466$ samples) of the clean dataset is required to completely remove our Universal Backdoor, which is more data than most victims can manually validate.

\section{Discussion and Related Work}
\textbf{Attacking web-scale datasets.}~\citet{carlini2023poisoning} demonstrate two realistic ways an attacker could poison a web-scale dataset: domain hijacking and snapshot poisoning.
They show that more than 0.15\% of the samples in these online datasets could be poisoned by an attacker.
However, existing many-to-one poison label attacks cannot exploit these vulnerabilities, as they require compromising many samples \emph{in a single class}~\citep{badnets, latent-seperation, blend-backdoor}.
As web-scale datasets contain thousands of classes~\citep{imagenet, openimages}, it is improbable that any one class would have enough compromised samples for a many-to-one poison label attack.
By leveraging inter-class poison transferability, our backdoor can utilize compromised samples outside a class the attacker is attempting to misclassify into.
\textbf{Scaling to larger datasets.}
The largest dataset we evaluate is our ImageNet-6K dataset, which consists of $6\,000$ classes and $7\,804\,447$ samples.
We created a Universal Backdoor in a model trained on this dataset while poisoning only 0.15\%.
As our backdoor effectively scales to datasets containing more classes and samples, we expect a smaller proportion of poison samples to be required to backdoor models trained on larger datasets, like LAION-5B~\citep{dataset-laion5b}.

\textbf{Alternative methodology for targeting multiple classes at inference time.}
Although we are the first to study how to target every class in the data poisoning setting, other types of attacks, like \emph{adversarial examples}, can be used to target specific classes at inference time~\citep{adv-survey, adv_examples}.
Through direct optimization on an input, the attacker finds an adversarial perturbation that acts as a trigger; adding it to the input causes a misclassification.
Defenses against adversarial examples seek to make models robust against adversarial perturbations~\citep{randomized-smoothing, adv_defense}.
However, as data poisoning backdoors utilize triggers that are not adversarial perturbations, these defenses are ineffective at mitigating data poisoning backdoors.

\textbf{Limitations.} \revised{ We focus on patch and blend triggers that are visible modifications to the image and hence could be detected by a data sanitation defense.
Our attacks are agnostic to the trigger; even if a specific trigger could be reliably detected, universal backdoors remain a threat because the attacker could have used a different trigger. 
\cite{data_san} demonstrate that no detection has been shown effective against any trigger.
However, evading data sanitation comes at a cost for the attacker:  Less detectable triggers are less effective at equal numbers.
Hence, the attacker must inject more to create an equally effective backdoor~\citep{attack-strength}.
We point to \Cref{sec:data-sanitation} showing that our attacks still remain difficult to detect using STRIP~\citep{strip} due to the high false positive rate. 
We focus on the feasibility of universal attacks and do not study the detectability-effectiveness trade-off of triggers with our attacks.  
Moreover, we focus on poisoning models from scratch, as opposed to poisoning pre-trained models that are fine-tuned.
More research is needed to analyze the effectiveness of our attacks against large pre-trained models like ViT and CLIP~\citep{vit, CLIP} that are fine-tuned on poisoned data.
Finally, we assume that the attacker can access similarly accurate surrogate classifiers to generate latent encodings for our attacks.
}

\section{Conclusion}

We introduce Universal Backdoors, a data poisoning backdoor that targets every class.
We establish that our backdoor requires significantly fewer poison samples than independently attacking each class and can effectively attack web-scale datasets.
We also demonstrate how compromised samples in uncritical classes can be used to reinforce poisoning attacks against other more sensitive classes.
Our work exemplifies the need for practitioners who train models on untrusted data sources to protect the whole dataset, not individual classes, from data poisoning.
Finally, we show that existing defenses are ineffective at defending against Universal Backdoors, indicating the need for new defenses designed to remove backdoors that target many classes.

\bibliography{iclr2024_conference}
\bibliographystyle{iclr2024_conference}

\appendix
\section{Appendix}
\label{sec:appendix_defense_params}
\Cref{tab:defense_params} contains a complete summary of all the parameters used to evaluate defenses against our backdoor in \Cref{sec:defenses}.
All defense parameters are adapted from \cite{lukas2023_pick_your_poison}, where they were optimized against a BadNets~\citep{badnets} patch trigger.
When hyperparameter tuning for fine-tuning and fine-pruning defenses, we find no significant improvements over the settings described in \cite{lukas2023_pick_your_poison}.
We reduce the fine-pruning rate in Fine-pruning, as we find it degrades the model's clean accuracy below our 2\% cutoff.

\begin{table}[ht]
\centering
\caption{Defense Parameters on ImageNet-1K from \cite{lukas2023_pick_your_poison}.}\label{tab:defense_params}

\begin{minipage}{0.45\linewidth}
\centering
\begin{tabularx}{\linewidth}{lc}
\toprule
\multicolumn{2}{c}{\textbf{Neural Attention Distillation}} \\
\midrule
n steps / N & 8,000 \\
opt & sgd \\
lr / $\alpha$ & 5e-4 \\
teacher steps & 1,000 \\
power / p & 2 \\
at lambda / $\lambda_{at}$ & 1,000 \\
weight decay & 0 \\
batch size & 128 \\
\midrule
\multicolumn{2}{c}{\textbf{Neural Cleanse}} \\
\midrule
n steps / N & 3,000 \\
opt & sgd \\
lr / $\alpha$ & 5e-4 \\
steps per class / N1 & 200 \\
norm lambda / $\lambda_N$ & 1e-5 \\
weight decay & 0 \\
batch size & 128 \\
\bottomrule
\end{tabularx}
\end{minipage}
\hfill
\begin{minipage}{0.45\linewidth}
\centering
\begin{tabularx}{\linewidth}{lc}

\midrule
\multicolumn{2}{c}{\textbf{Fine-Tuning}} \\
\midrule
n steps / N & 5,000 \\
opt & sgd \\
lr / $\alpha$ & 5e-4 \\
weight decay & 0.001 \\
batch size & 128 \\
\midrule
\multicolumn{2}{c}{\textbf{Fine-Pruning}} \\
\midrule
n steps / N & 5,000 \\
opt & sgd \\
lr / $\alpha$ & 5e-4 \\
prune rate / $\rho$ & 10\% \\
sampled batches & 10 \\
weight decay & 0 \\
batch size & 128 \\
\bottomrule
\end{tabularx}

\end{minipage}

\end{table}

\begin{figure}
    \begin{subfigure}{0.515\textwidth}
        \includegraphics[width=\linewidth]{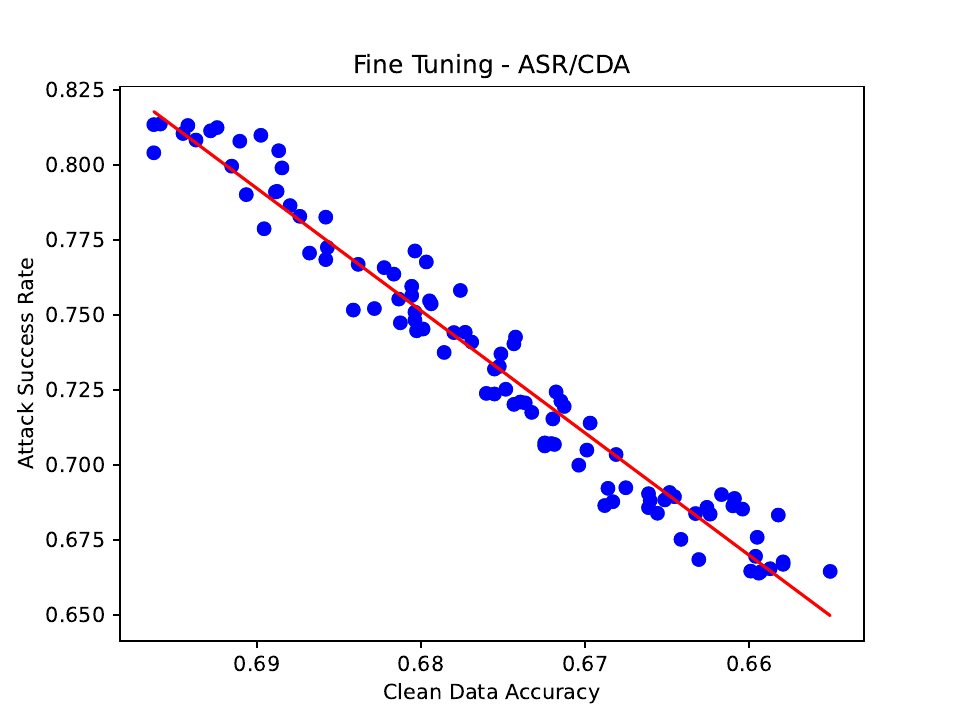}
        \caption{}
        \label{fig:apx_tradeoff}
    \end{subfigure}%
    \begin{subfigure}{0.485\textwidth}
\includegraphics[width=\linewidth]{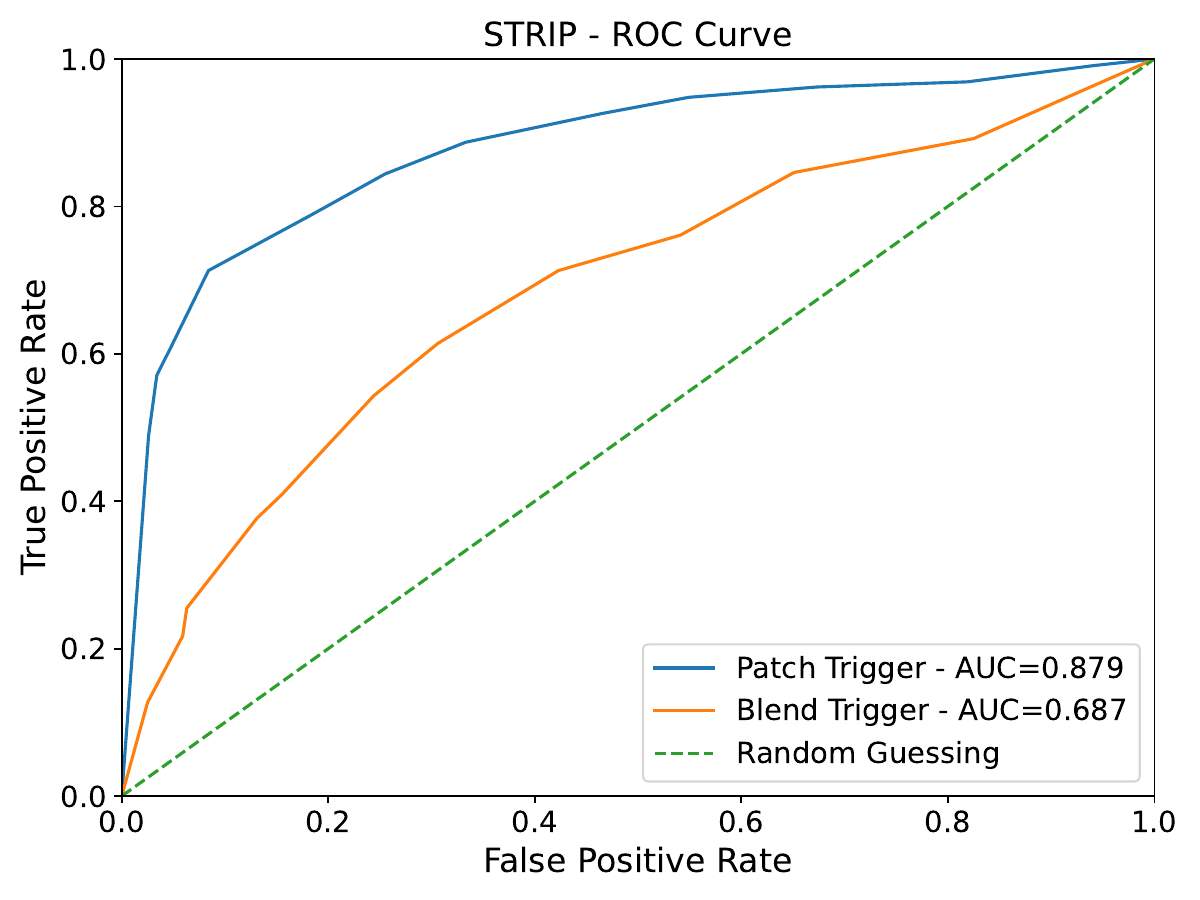}
        \caption{}
        \label{fig:roc}
    \end{subfigure}
    \caption{(\labelcref{fig:apx_tradeoff}) Trade-off between attack success rate and clean data accuracy when fine-tuning a backdoored model. (\labelcref{fig:roc}) ROC curve of our Universal Backdoor with patch and blend triggers (see \Cref{fig:example-triggers}) when applying the STRIP~\citep{strip} defense.}
\end{figure}

\revised{As shown by \Cref{fig:apx_tradeoff}, a linear trade-off exists between the effectiveness of defenses and the allowed clean accuracy cutoff.
If the defender allows for more clean accuracy degradation, the effectiveness of the backdoor can be further reduced.
This does not apply to all defenses, as defenses like neural cleanse~\citep{neural-cleanse} do not significantly reduce clean accuracy.}

\subsection{Analysis of Clean Accuracy}
\label{sec:apx-clean-acc}

\revised{If a backdoor attack degrades the clean accuracy of a model, then the validation set is sufficient for the victim to recognize the presence of a backdoor~\citep{badnets}.
Therefore, a model trained on the poisoned set should achieve the same clean accuracy as one trained on a comparable clean dataset.
We find that our backdoored models have the same clean accuracy across all runs as a model trained on entirely clean data.
We train a clean ResNet-18 model on ImageNet-1k~\citep{imagenet1k}, which achieves 68.49\% top-1 accuracy on the validation set.
\Cref{tab:apx-clean} shows the clean accuracy of backdoored models on the ImageNet-1k dataset.}

\begin{table}[ht]
\centering
\caption{Clean accuracy of backdoored models on ImageNet-1k dataset.}
\label{tab:apx-clean}
\begin{tabular}{c|c|cc|cc}
\toprule
Poison Samples ($p$) & Poison \% & \multicolumn{2}{c|}{Patch} & \multicolumn{2}{c}{Blend} \\
\midrule
 & & Ours & Baseline & Ours & Baseline \\
\midrule
$2000$ & $0.16$ & 68.94\% & 68.94\% & 68.51\% & 69.43\% \\
$5000$ & $0.39$ & 68.92\% & 68.89\% & 68.77\% & 68.66\% \\
$8000$ & $0.62$ & 68.91\% & 69.43\% & 69.78\% & 69.22\% \\
\bottomrule
\end{tabular}
\end{table}

\subsection{Inter-class Poison Transferability With Small Variation Sets}
\label{sec:apx-transfer-classes}

\begin{table}[ht]
\centering
\caption{Effect of the number of classes in the variation set \textbf{A} on attack success on the observed set \textbf{B}. All experiments use 4\,600 poison samples.}
\label{tab:apx-transfer}
\begin{tabular}{c|c}
\toprule
Percentage of classes in \textbf{A} & ASR on classes in \textbf{B} \\
\midrule
90\% & 72.77\%  \\
60\% & 70.45\%  \\
30\% & 71.78\%  \\
10\% & 67.72\%  \\
\bottomrule
\end{tabular}
\end{table}

\revised{\Cref{tab:apx-transfer} shows that even if the number of classes in the variation set \textbf{A} is reduced to only 10\% of classes in $\mathcal{Y}$, inter-class poison transferability maintains its effect on the observed set \textbf{B}.
This results in an otherwise unsuccessful attack on classes in \textbf{B}, achieving a success rate of 67.72\%.
Therefore, if the attacker can strongly poison a small set of classes in the dataset, attacking other classes in the model can easily be accomplished, as inter-class poison transferability remains strong.
To protect even a tiny subset of high-value classes, the victim must maintain the integrity of every class within their dataset.}

\subsection{Data Sanitation Defenses}
\label{sec:data-sanitation}
Several data sanitization defenses are also poorly suited to Universal Backdoors.
SPECTRE~\citep{data_sanitation_4} only removes samples from a single class by design, and therefore could remove at most 0.1\% of our Universal Backdoor's poisoned samples on ImageNets-1K. 
STRIP~\citep{strip} struggles to detect our trigger, resulting in a high false positive rate, as shown in \Cref{fig:roc}.
The area under the ROC curves are $0.879$ and $0.687$ for the patch and blend triggers, respectively. 
It may be difficult for defenders to detect both triggers for large datasets (1 million samples or more) due to the detection's high FPR. 
Considering a maximum tolerable FPR of $10\%$, the defender misses $39\%$ of the patch trigger samples and $68\%$ of the blended triggers.

\subsection{Class-wise Attack Success Metrics}
\label{sec:apx-attack-per-class}
\revised{Our method does not achieve even attack success across all classes in the dataset.
\Cref{tab: apx-metrics-table} shows statistics of our Universal Backdoor's success rate across classes in ImageNet-1K.
We find that some classes are more challenging to achieve a successful attack against our backdoor.
This differs from the baseline, as the baseline either performs near-perfectly or not at all.}

\begin{table}[h]
\centering
\caption{ASR metrics across classes in ImageNet-1K}
\begin{tabular}{ccccc}
\toprule
Poison Samples (p) & Min & Max (\%) & Mean & Median \\ 
\midrule
2000                        & 0                 & 100\%               & 81.0\%                 & 98.0\%                   \\ 
5000                        & 0                 & 100\%               & 95.4\%               & 100\%                  \\ 
\bottomrule
\end{tabular}
\label{tab: apx-metrics-table}
\end{table}

\end{document}

%% file: math_commands.tex
\usepackage{amsmath,amsfonts,bm}

\def\eqref#1{equation~\ref{#1}}

\def\1{\bm{1}}

\def\vb{{\bm{b}}}
\def\vc{{\bm{c}}}

\def\vx{{\bm{x}}}

\DeclareMathAlphabet{\mathsfit}{\encodingdefault}{\sfdefault}{m}{sl}
\SetMathAlphabet{\mathsfit}{bold}{\encodingdefault}{\sfdefault}{bx}{n}

